\documentclass[preprint,review,10pt]{elsarticle}
 
\usepackage{graphicx} 
\usepackage{amsmath}
\usepackage{amsfonts}
\usepackage{xcolor}
\usepackage{amssymb}
\usepackage{caption}
\usepackage{float}
\usepackage{algorithm}
\usepackage{algorithmic}
\usepackage{tabularray}
\usepackage{rotating}

\usepackage{amssymb}
\usepackage{amsmath}

\newcommand\restr[2]{\ensuremath{\left.#1\right|_{#2}}}

\begin{document}

\begin{frontmatter}



\title{Hybrid machine learning based scale bridging framework for permeability prediction of fibrous structures}


\author [label1] {Denis Korolev} 
\author [label2] {Tim Schmidt}
\author [label3] {Dinesh K. Natarajan}
\author [label2] {Stefano Cassola}
\author [label4] {David May}
\author [label2] {Miro Duhovic}
\author [label1,label5] {Michael Hintermüller}

\affiliation[label1]{organization={Weierstrass-Institute for Applied Analysis and Stochastics},
            addressline={Mohrenstr. 39}, 
            city={Berlin},
            postcode={10117}, 
            country={Germany}}

\affiliation[label2]{organization={Leibniz-Institut für Verbundwerkstoffe GmbH},
            addressline={Erwin-Schrödinger-Str. 58}, 
            city={Kaiserslautern},
            postcode={67663}, 
            country={Germany}}
            
\affiliation[label3]{organization={German Research Center for Artificial Intelligence GmbH (DFKI)},
            addressline={Trippstadter Str. 122}, 
            city={Kaiserslautern},
            postcode={67663}, 
            country={Germany}}
\affiliation[label4]{organization={Faserinstitut Bremen e.V. (FIBRE) University of Bremen},
            addressline={Am Biologischen Garten 2, Geb. IW 3}, 
            city={Bremen},
            postcode={28359}, 
            country={Germany}}

\affiliation[label5]{organization={Institute for Mathematics, Humbolt-Universität zu Berlin},
            addressline={Unter den Linden 6}, 
            city={Berlin},
            postcode={10099}, 
            country={Germany}}
            
\begin{abstract}

This study introduces a hybrid machine learning-based scale-bridging framework for predicting the permeability of fibrous textile structures. By addressing the computational challenges inherent to multiscale modeling, the proposed approach evaluates the efficiency and accuracy of different scale-bridging methodologies combining traditional surrogate models and even integrating physics-informed neural networks (PINNs) with numerical solvers, enabling accurate permeability predictions across micro- and mesoscales. Four methodologies were evaluated: fully resolved models (FRM), numerical upscaling method (NUM), scale-bridging method using data-driven machine learning methods (SBM)  and a hybrid dual-scale solver incorporating PINNs (PINN). The FRM provides the highest fidelity model by fully resolving the micro- and mesoscale structural geometries, but requires high computational effort. NUM is a fully numerical dual-scale approach that considers uniform microscale permeability but neglects the microscale structural variability. The SBM accounts for the variability through a segment-wise assigned microscale permeability, which is determined using the data-driven ML method.  This method shows no significant improvements over NUM with roughly the same computational efficiency and modeling runtimes of ~45 minutes per simulation. The newly developed  hybrid dual-scale solver incorporating PINNs shows the potential to overcome the problem of data scarcity of the data-driven surrogate approaches. The hybrid framework advances permeability modeling by balancing computational cost and prediction reliability, laying the foundation for further applications in fibrous composite manufacturing, but is not yet mature enough to be applied to large complex geometry models.
\end{abstract}


\begin{highlights}
\item Enhanced methodologies for permeability prediction of textile-based fiber structures

\item Evaluation of four different methods for determining the mesoscale permeability of 3D models

\item Dual-scale permeability prediction framework with a novel hybrid physics-informed dual-scale solver

\item Application of the hybrid physics-informed dual-scale solver to microscale flow approximation for permeability prediction of 2D fiber structures

\end{highlights}


\end{frontmatter}



\section{Introduction}

\begin{figure}[H]
    \centering
    \includegraphics[width=1\linewidth]{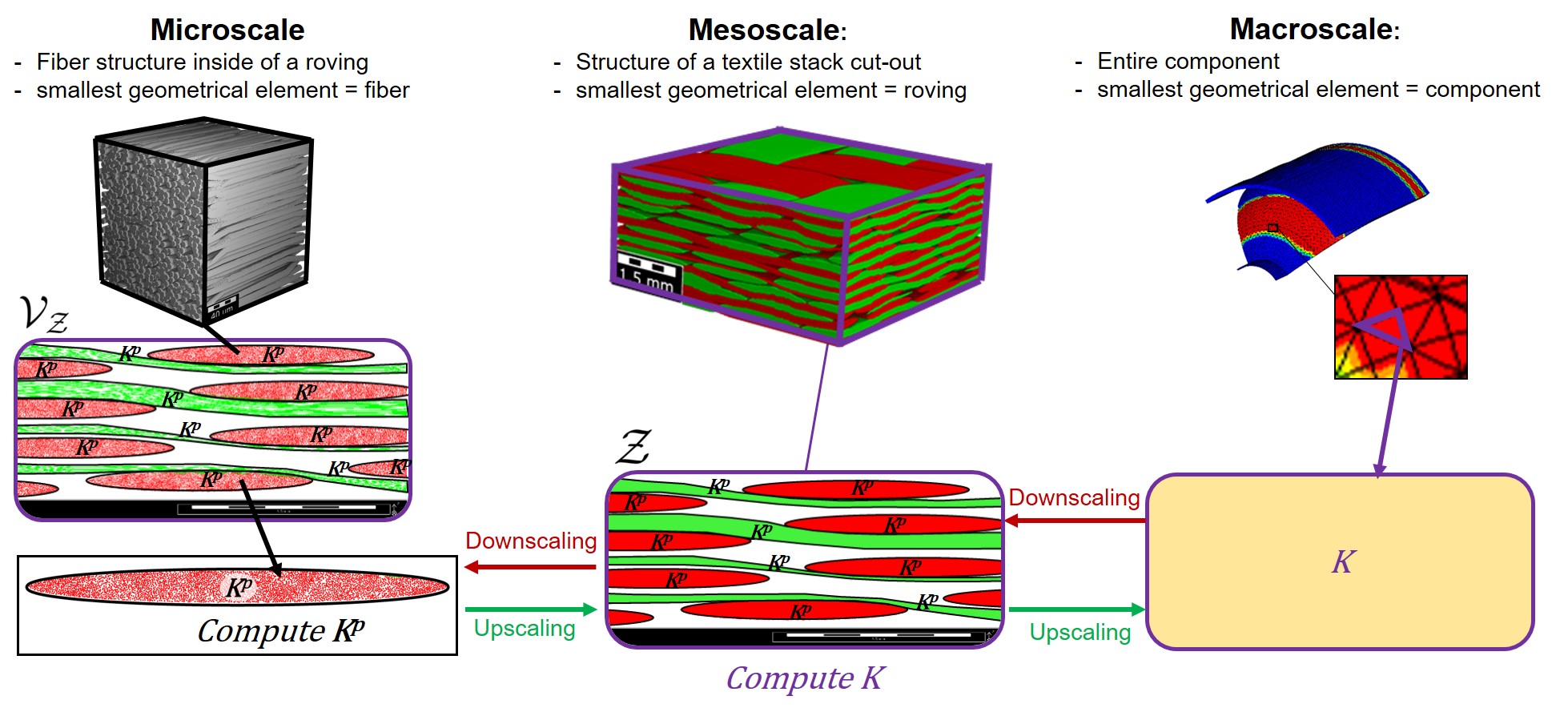}
    \caption{Relevant scale levels for permeability prediction.} 
    \label{fig:enter-label1}
\end{figure}

The manufacturing of fiber-reinforced polymers (FRP) entails combining a fiber reinforcement structure with a matrix polymer, typically by infiltrating the fibers with a resin system driven by a pressure difference. Predicting resin flow within this structure requires considering multiple scales: on the microscale, flow between impermeable fibers is simulated; on the mesoscale, flow between permeable fiber bundles, so-called rovings, becomes relevant; and on the macroscale, mold filling is considered \cite{lomov2001textile}. Since simulating multiple scales simultaneously requires immense computing resources, they are typically treated separately. However, data must be exchanged between scales: structural information from meso- to microscale (referred to as downscaling) and, conversely, the assignment of microscale permeability to the permeable rovings at the mesoscale (referred to as upscaling), as illustrated in Fig.\ref{fig:enter-label1}. This combination of down- and upscaling , referred to as scale bridging \cite{lubbers2020modeling}, requires the homogenization of data for the exchange between scales. In addition to the homogenization, scale separation and simulation lead to simplifications in the underlying physics \cite{gorguluarslan2014simulation}. These simplifications, however, are necessary to reduce the computational costs. Still, state-of-the-art permeability determination methods require multiple computationally expensive simulations, and assigning permeability to rovings at the higher scale is complex due to significant local variations in the textile and fiber structure, such as fiber orientation and volume fraction, which increases the effort \cite{seuffert2021micro}.

Scientific machine learning (SciML) appears capable of aiding the scale-bridging process, but its success depends on overcoming a challenge that most surrogate modeling approaches struggle with: effectively bridging scales while incorporating physical laws, data-driven insights from experimental observations, and scale relations into a unified framework. Our approach proposes several scale-bridging methods for computing permeability, supported by supervised deep learning to replace costly numerical simulations. While these methods offer significant speedups, data-driven approaches inherently suffer from limitations such as the scarcity of training data. This can lead to prediction errors on unseen data (generalization errors) and biases toward the training data (overfitting), ultimately reducing the overall reliability of these methods.

To address the limitations of data-driven approaches, we propose a complementary physics-informed machine learning method that reduces reliance on data and the associated microscale solver by incorporating physical knowledge, such as governing equations and structural properties of the multiscale system. Following \cite{raissi2019physics}, we reformulate the microscale PDE solver as a physics-informed neural network (PINN) optimization problem. Building on the multiscale framework of \cite{hintermuller2023hybrid}, we enhance the PINN approach by exploiting the dual-scale structure of permeability computation. Specifically, we couple meshless PINNs used for microscale flow approximation within fiber tows with a robust mesh-based mesoscale numerical solver to approximate the flow between the tows, where meshing is generally easier. Both solvers are integrated into a monolithic dual-scale framework via scale bridging. This setup enables the integration of data from both scales, including surrogate-derived permeability values, into a multi-fidelity framework that improves permeability predictions by resolving microscale flow without relying on mesh-dependent microscale numerical solvers. The dual-scale solver supports the robust application of periodic boundary conditions, which are challenging to impose using standard PINNs in fiber geometries, and generally increases the reliability of the PINN approach by mitigating poor microscale approximation through coarse-scale regularization.



\section{Upscaling and Downscaling}
\label{Section: Upscaling and Downscaling}

For the textile stack model $\Omega_{\text{Me}} \subset \mathbb{R}^{d}$ and its microscopic representation $\Omega_{\text{Mi}}\subset \mathbb{R}^{d}$, let $\mathcal{M}_{\text{Mi}}=\{ \mathcal{V}   \mid  \mathcal{V} \subseteq \Omega_{\text{Mi}} \}$ and $\mathcal{M}_{\text{Me}}=\{\mathcal{Z}   \mid  \mathcal{Z} \subseteq \Omega_{\text{Me}} \}$ denote the sets of  micro- and mesoscale geometry models, respectively. In mesomodels $\mathcal{Z} \in \mathcal{M}_{\text{Me}}$, represented as mesoscale statistical volume elements (SVE) in the form of textile stack cut-outs, porous rovings are modeled as a continuum, and the SVEs are divided into fluid parts $\mathcal{Z}^{F}$ and porous parts $\mathcal{Z}^{P}$, such that $\mathcal{Z} = \mathcal{Z}^{P} \cup \mathcal{Z}^{F}$. The resolution operator
\begin{align}
\mathcal{F}_{\downarrow}:  \ \mathcal{M}_{\text{Me}} \rightarrow \mathcal{M}_{\text{Mi}} \quad \text{with} \quad  \mathcal{Z}  \mapsto \mathcal{V}_{\mathcal{Z}}:=\mathcal{F}_{\downarrow}(\mathcal{Z}), 
\end{align}
transfers mesomodels into their respective micromodels $\mathcal{V}_{\mathcal{Z}} \in \mathcal{M}_{\text{Mi}}$, corresponding to fully resolved mesoscale SVEs, as in Fig. \ref{fig:enter-label1}. The micromodels are divided into fluid $\mathcal{V}_{\mathcal{Z}}^{F}$ and solid $\mathcal{V}_{\mathcal{Z}}^{S}$ parts. The microstructures $\mathcal{V}_{\mathcal{Z}}^{S}$ (bundles of impermeable fibers) are derived from the mesostructure of $\mathcal{Z}$ including the local orientation of the rovings and the fiber volume content (FVC) in the rovings. With these features and the fiber diameter, suitable microscale SVEs as statistical approximations of resolved mesomodels can be generated as well, reducing simulation costs. This transfer of information from the mesoscale to the microscale is referred to as \textbf{downscaling}.

The flow of information from the micro- to the mesoscale is called \textbf{upscaling}. For its realization, we define the upscaling operator 
\begin{align}
\mathcal{F}_{\uparrow}: \  \mathcal{M}_{\text{Mi}} \rightarrow \mathbb{R}^{d \times d} \quad \text{with} \quad  \mathcal{V}_{\mathcal{Z}} \mapsto  \boldsymbol{K}^{p}[\mathcal{V}_{\mathcal{Z}}],
\end{align}
which extracts the geometric information about the microstructures of $\mathcal{V}_{\mathcal{Z}}$ in the form of the permeability tensor $\boldsymbol{K}^{p}[\mathcal{V}_{\mathcal{Z}}]$ (also referred to as micropermeability) to characterize the porous part $\mathcal{Z}^{P}$ of $\mathcal{Z}$. Computing $\boldsymbol{K}^{p}[\mathcal{V}_{\mathcal{Z}}]$ requires solving the algebraic system derived from Darcy's law:
\begin{align}\label{Darcy's law}
\boldsymbol{U} = \frac{1}{\mu}\boldsymbol{K}^{p}[\mathcal{V}_{\mathcal{Z}}] \boldsymbol{PD},
\end{align}
where $\mu$ is the dynamic viscosity of the fluid, $\boldsymbol{U} \in \mathbb{R}^{d \times d}$ and $\boldsymbol{PD} \in \mathbb{R}^{d \times d}$ are the volume-averaged flow velocity and pressure drop matrices with
\begin{align}\label{averaging process}
\begin{aligned}
\boldsymbol{U}_{k,j} =  \frac{1}{|\mathcal{V}_{\mathcal{Z}}^{F}|}\int_{\mathcal{V}_{ \mathcal{Z}}^{F}} u_{j}^{(k)} (x) \  dx, \quad 
\boldsymbol{PD}_{k,j}  = \boldsymbol{f}_{j}^{(k)} - \frac{1}{|\mathcal{V}_{\mathcal{Z}}^{F}|} \int_{\mathcal{V}_{\mathcal{Z}}^{F}} \frac{\partial p^{(k)}}{\partial x_{j}}(x)  \  dx.
\end{aligned}
\end{align}
The respective velocity fields $\boldsymbol{u}^{(k)} = [u_{1}^{(k)},\  u_{2}^{(k)}, \  u_{3}^{(k)}]$ and pressures $p^{(k)}$ are obtained by solving three Stokes equations (for $d=3$) at the microscale: 
\begin{equation}\label{Stokes equation}
\begin{aligned}
- \mu \Delta \boldsymbol{u}^{(k)} + \nabla p^{(k)} &= \boldsymbol{f}^{(k)} \quad &\text{in} \ \mathcal{V}_{\mathcal{Z}}^{F}, \\
\nabla \cdot \boldsymbol{u}^{(k)} &= 0 \quad &\text{in} \ \mathcal{V}_{\mathcal{Z}}^{F}, \\ 
\boldsymbol{u}^{(k)} &= 0 \quad  &\text{on} \ \partial \mathcal{V}_{\mathcal{Z}}^{S},
\end{aligned}
\end{equation}
where $\partial \mathcal{V}_{\mathcal{Z}}^{S}$ denotes the (no-slip) boundary of $\mathcal{V}_{ \mathcal{Z}}^{S}$ and $\boldsymbol{f}^{(k)}$ is a volume force. For example, one applies $\boldsymbol{f}^{(k)}=\boldsymbol{e}^{k}$, where $\boldsymbol{e}^{k}$ is the unit vector (unit pressure drop) in $k$-th direction, and uses periodic boundary conditions for $\boldsymbol{u}^{(k)}$ and $p^{(k)}$ to get such three solutions; cf. \cite{griebel2010homogenization} and references therein.  In this case, the second term contributing to $\boldsymbol{PD}_{k,j}$ in \eqref{averaging process} vanishes, and the permeability becomes a function of the average velocity field. Solving \eqref{Stokes equation} on a fully resolved model $\mathcal{V}_{\mathcal{Z}}$ of the mesoscale geometry $\mathcal{Z}$ is computationally expensive or even prohibitive. Therefore, the SVE approximation of $\mathcal{V}_{\mathcal{Z}}$ is often used instead. 

At the mesoscale, the flow is modeled by the Stokes-Brinkman equation:
\begin{equation}\label{Stokes-Brinkman flow}
\begin{aligned}
- \tilde{\mu} \Delta \boldsymbol{u}^{SB 
 (k)} + \mu \big(\boldsymbol{K}_{\text{Me}}[\mathcal{V}_{\mathcal{Z}}]\big)^{-1} \boldsymbol{u}^{SB (k)}  + \nabla p^{SB (k)} &=  \boldsymbol{f}^{(k)}  \quad \text{in} \ \mathcal{Z}, \\ 
\nabla \cdot \boldsymbol{u}^{SB  (k)}  & =  0   \quad \quad \ \text{in} \ \mathcal{Z}, 
\end{aligned}
\end{equation}
where $\tilde{\mu}$ is the effective Brinkman viscosity, and $\boldsymbol{K}_{\text{Me}}[\mathcal{V}_{\mathcal{Z}}] = \chi_{\mathcal{Z}^{F}} \cdot \infty + \chi_{\mathcal{Z}^{P}} \cdot \boldsymbol{K}^p[\mathcal{V}_{\mathcal{Z}}]$ ($\chi_{\mathcal{Z}^{F}}$ and $\chi_{\mathcal{Z}^{P}}$ are the indicator functions of $\mathcal{Z}^{F}$ and $\mathcal{Z}^{P}$) represents the permeability properties of different domain regions. Three solutions of \eqref{Stokes-Brinkman flow} must be computed to determine the mesoscale permeability $\boldsymbol{K}[\mathcal{Z}]$ using \eqref{Darcy's law} while averaging the mesoscale variables over $\mathcal{Z}$.

\section{Numerical methods for dual-scale permeability prediction}
\label{Section: Numerical methods for dual scale permeability predcition}

According to the state of the art, the micro- and mesoscale are modeled and simulated separately in order to numerically determine the dual-scale permeability of a textile stack \cite{schmidt2019novel, nedanov2002numerical, syerko2017numerical}. For this, geometry models are required that represent the fiber structure. To ensure that the models account for all relevant structural properties, including realistic variations, studies are conducted in advance with the aim of determining SVEs for the fiber structure \cite{du2006size, Schmidt2025}. In addition to the structural features, suitable model resolution, model size, and the number of models must be identified during SVE development to achieve a distribution of permeabilities similar to the experiments \cite{syerko2023benchmark}.

\begin{figure}[H]
    \centering
    \includegraphics[width=1\linewidth]{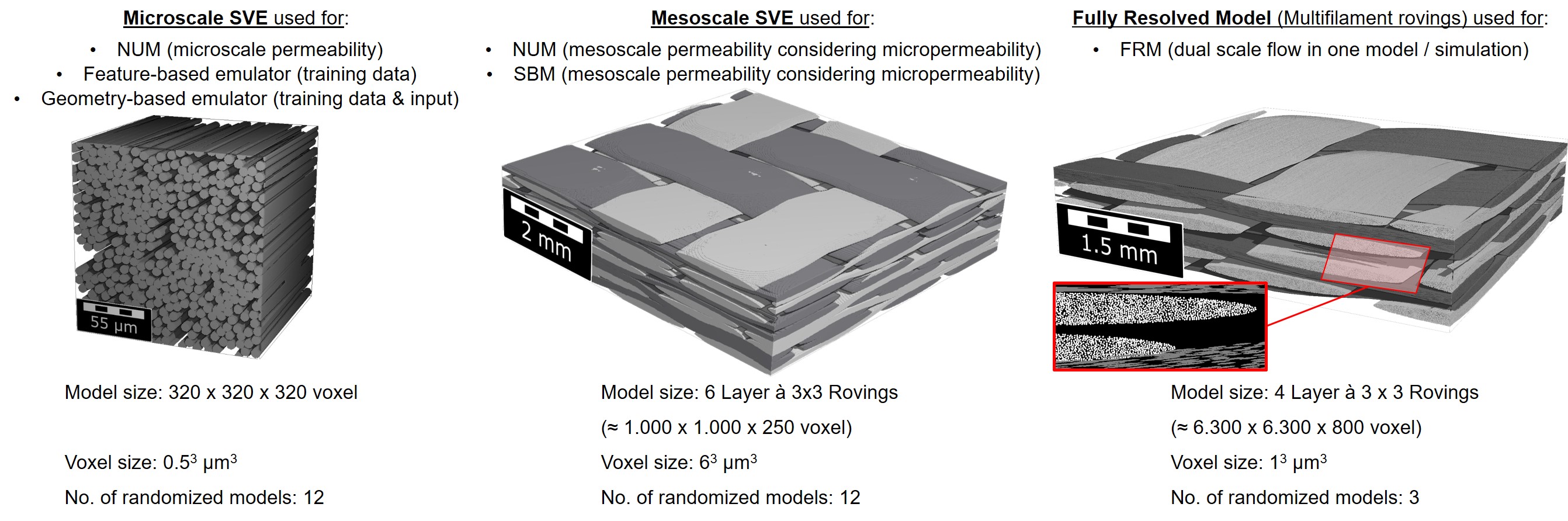}
    \caption{Visualisation of the statistical representative volume elements for micro- and mesoscale and the fully resolved model}
    \label{fig:enter-label2}
\end{figure}

The 3D models were generated in the GeoDict\textsuperscript{\textregistered} software and are voxel-based, with voxels (equilateral hexahedra akin to image pixels) classified as solid, fluid, or porous. The microscale SVEs contains several hundred fibers \cite{Schmidt2025}, and the mesoscale SVEs contains several textile layers, which are generated using rovings, see Fig. \ref{fig:enter-label2}. The rovings are modeled as a continuum, which allows the use of a coarser resolution. To create the model, individual textile layers with varying roving cross-sections, roving undulation and distances between rovings are generated and positioned randomly on top of each other within a defined range, followed by a virtual compaction of the textile stack until the desired stack height or FVC is achieved. This causes the rovings to deform and the textile layers to nest into each other, as in reality. To account for the flow within the rovings, the anisotropic micropermeability must be assigned. In GeoDict, each voxel has a material ID that allows properties such as anisotropic permeability to be defined. All methods are shown in Fig. \ref{fig:enter-label3}. All numerical flow simulations were carried out in the GeoDict module FlowDict. For microscale models and fully resolved models (FRM) the LIR solver was used for the Stokes problem \eqref{Stokes equation}. For the numerical upscaling method (NUM) and the scale bridging method (SBM) the SimpleFFT solver was used for the Stokes-Brinkman problem \eqref{Stokes-Brinkman flow} \cite{flowdict2024user}. Periodic boundary conditions were applied for all simulations in tangential and in flow direction, with additional empty inflow and outflow regions of 40 voxels. The fiber-free inflow and outflow regions allow the flow to enter and exit the fiber structures of the model without obstruction. An error bound of 0.01, corresponding to a change in permeability of less than 1\%, was chosen as the stopping criterion \cite{flowdict2024user}.

\subsection{Numerical Upscaling Method - NUM}
\label{sec:sum}

For NUM and SBM, 12 individual models for each of five compaction levels (60 models) with a voxel resolution of \((6^{3} \ \mu\mathrm{m}^{3}/\text{voxel})\) and six textile layers were generated. For the NUM, all warp and weft rovings have the same material ID and, therefore, the same micropermeability tensor. In the roving direction, assumed to align with the respective model axis, the rovings are assigned the fiber-direction micropermeability. Perpendicular directions are assigned the transverse micropermeability. This implies that local variations within the roving structure, such as those induced by the weave structure or deformations, are neglected.

\subsection{Fully resolved Models - FRM}

The scales are not separated in the FRM and therefore no microscale homogenization takes place. A small voxel size must be chosen for the entire model to resolve the flow between the fibers \((1^{3} \ \mu\mathrm{m}^{3}/\text{voxel})\). This results in large geometry models and flow simulations, which can only be computed on high-performance clusters. The model generation procedure was comparable to the NUM and SBM approaches, with the only difference being that multifilament rovings (3000 fibers with a diameter of 7 µm) were created instead of rovings modeled as a continuum. However, using an appropriate resolution \((0.5^{3} \ \mu\mathrm{m}^{3}/\text{voxel})\) for accurate microscale representation at a mesoscale SVE size of 6 layers with 3x3 rovings per layer exceeded 2 TB of RAM during the flow simulation, which was the largest available memory. As a compromise between size and resolution, three models for six compaction levels were created with four layer and 2.4 x 2.4 rovings per layer and a voxel size of \(1^{3} \ \mu\mathrm{m}^{3}\)(5.100 x 5.100 x 716 - 1119 voxel).

\section{Data-driven methods for dual-scale permeability prediction (Scale-Bridging Method - SBM)}
\label{Section: Methods for dual scale permeability prediction}

\begin{figure}
    \centering
    \includegraphics[width=1\linewidth]{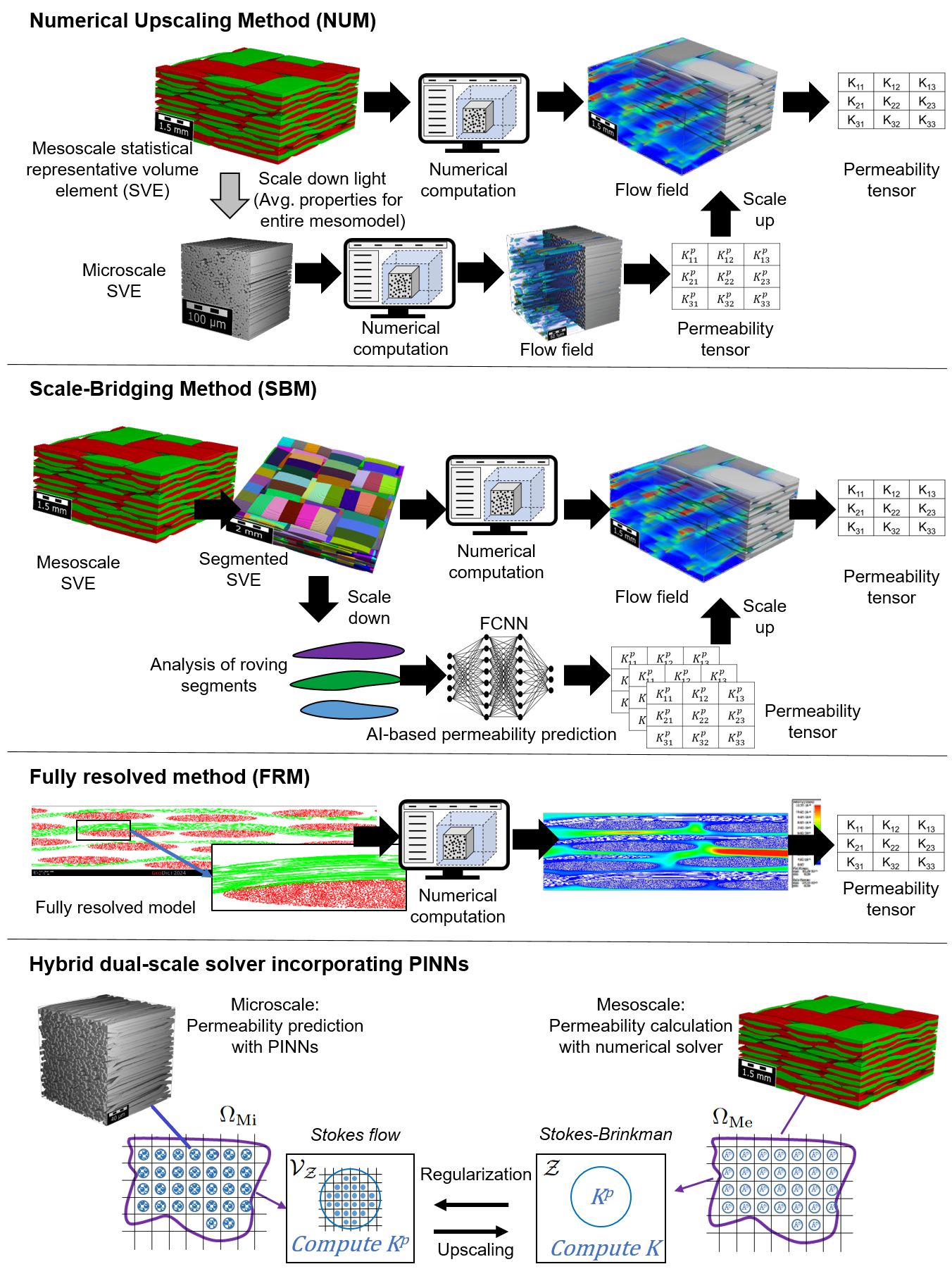}
    \caption{Schematic illustrations of all four dual-scale methods used for permeability prediction}
    \label{fig:enter-label3}
\end{figure}

The SBM approach was developed to reduce the inaccuracies of the NUM approach. In this method, the rovings are decomposed into several segments, see Fig. \ref{fig:enter-label3}, each with individual material IDs. Downscaling is performed for each segment: the FVC is caluclated using the cross-sections of the rovings, the fiber diameter and the number of fibers per roving, and the orientation of the roving segment, which approximately correspond to the fiber orientation within the roving, are determined. However, this method also represents a simplification of the local structure, as homogeneous properties are assumed for each segment. Therefore, the segment size affects the modeling accuracy. Based on the structural parameters: FVC, fiber orientation, and fiber diameter, the micropermeability is determined using the data-driven methods described in Section 4.1 and subsequently assigned to the material ID of the corresponding roving segment. This method can be used to assign up to 255 different anisotropic permeability values (due to the limited number of material IDs in GeoDict), resulting in more accurate micropermeabilities in the mesomodel compared to NUM.

\subsection{Feature-based and geometry-based emulator for permeability prediction}
\begin{figure}[H]
    \centering
    \includegraphics[width=0.9\linewidth]{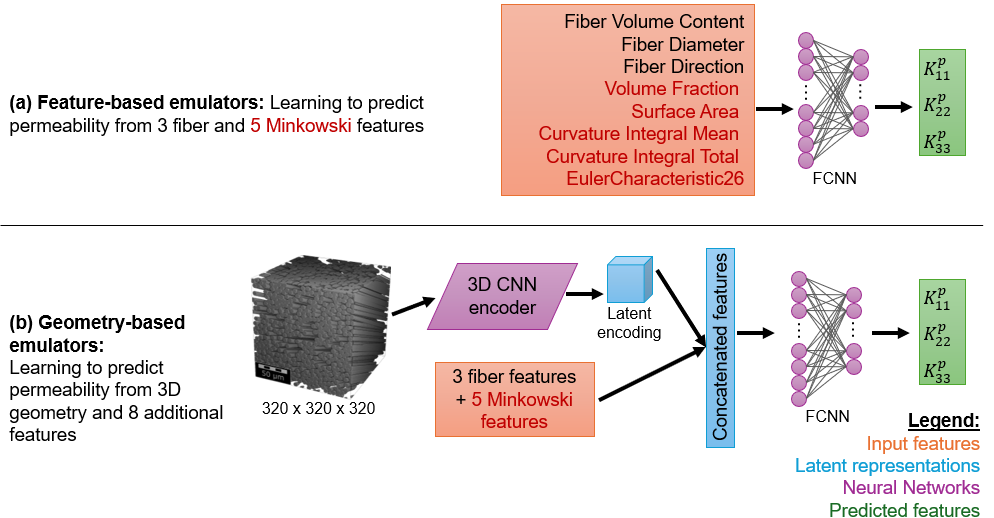}
    \caption{Two classes of modeling approaches for the ML emulators for permeability prediction. The approaches differ based on the input features used to predict the permeabilities. The emulators are constructed using fully connected neural networks and 3D convolutional neural networks. The architecture of the 3D CNN encoder is based on \cite{elmorsy_generalizable_2022}.}
    \label{fig:two-dd-approaches}
\end{figure}

Several recent studies have proposed ML methods as surrogate models or emulators for the permeability prediction of microstructures, i.e., on a single scale. Based on the input features used for prediction, these methods can be categorized into two classes: (1) feature-based methods and (2) geometry-based method, as shown in Fig. \ref{fig:two-dd-approaches}. Various modeling approaches for micropermeability prediction have been investigated in \cite{Schmidt2025, natarajan_data-driven_2024} and references therein. 

The main advantage of such ML emulators is their significant speed-up in inference times compared to numerical simulations, albeit with a trade-off in accuracy. \cite{natarajan_data-driven_2024} showed that feature-based emulators achieved an inference speed-up of $10^6$ with a relative error of $11.35\%$, whereas geometry-based emulators achieved an inference speed-up of $10^4$ with a relative error of $8.33\%$. Depending on the available features describing the microstructure, either feature-based or geometry-based emulators can be employed for micropermeability prediction.

\section{Physics-informed methods for dual-scale permeability prediction}

Our physics-based approaches for permeability prediction use PINNs as a backbone of the framework, as they offer several key advantages. Unlike traditional simulation methods that require mesh-based solvers, PINNs are meshless and solve PDEs through an optimization process that forces the underlying neural network ansatz to satisfy the (non-discretized) microscale equation \eqref{Stokes equation} at collocation points sampled from fibrous geometries. This can yield more accurate microscale flow approximations within fiber tows, where discretization in standard methods may affect permeability computation, while eliminating the need for complex meshing. PINNs support efficient distributed GPU implementation due to their excellent parallelization capabilities \cite{stiller2020large, escapil2023h} and domain decomposition techniques \cite{shukla2021parallel, moseley2023finite}, making them promising for demanding computational tasks such as upscaling, once such scalable implementations are transferred to GPU clusters. However, these scaling techniques, in the context of applications to complex 3D geometries, are particularly non-trivial. Thus, we focus on 2D periodic microscale and mesoscale geometries only. 

\subsection{Benchmark models for physics-informed approaches}
\label{Section: Benchmark models for physics-informed approaches}

\begin{figure}[H]
    \centering
    \includegraphics[width=1\linewidth]{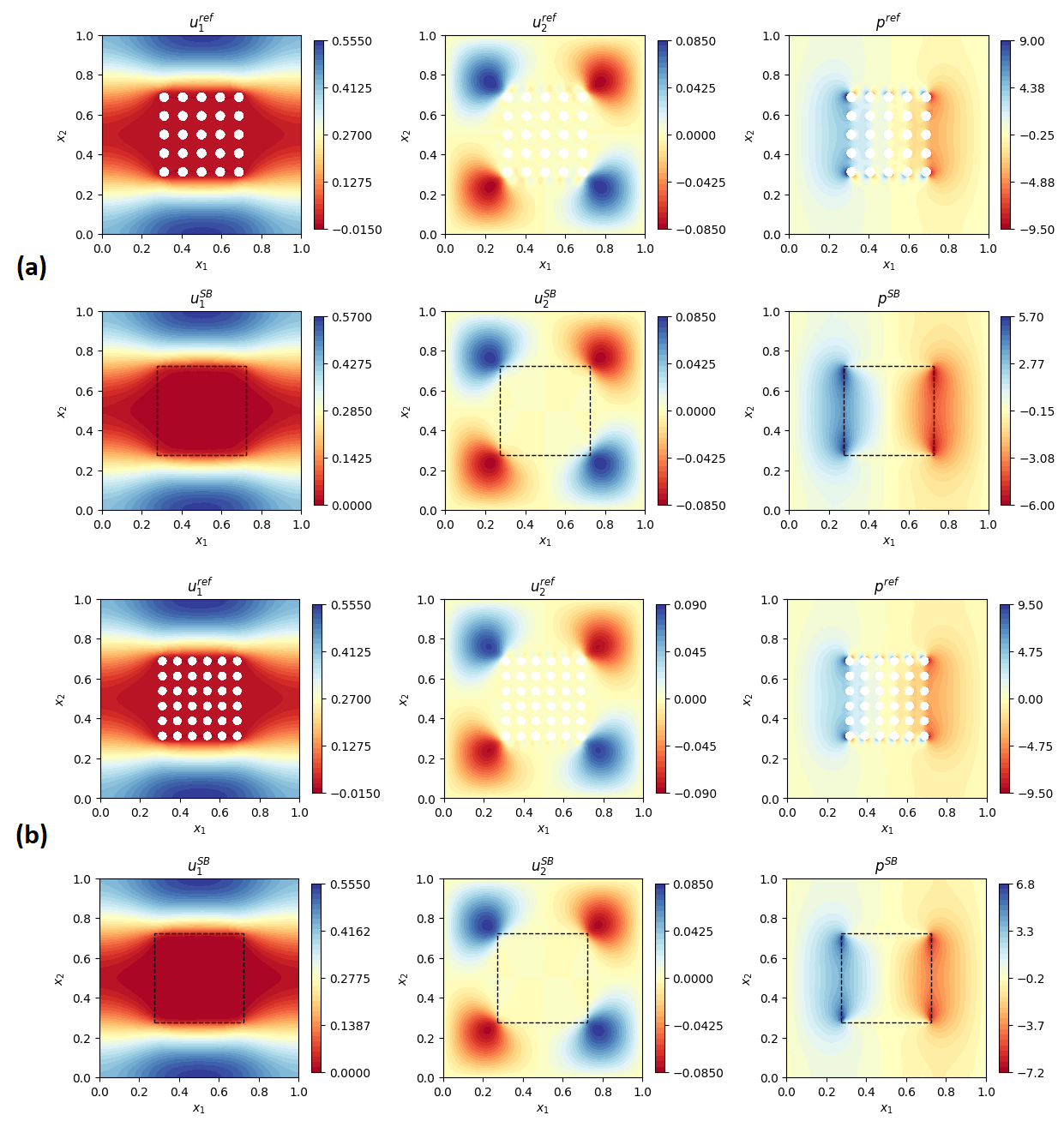}
    \caption{Two-scale 2D reference models: microscale $\mathcal{V}_{\mathcal{Z}_{i}}$ and mesoscale $\mathcal{Z}_{i}$ geometries and the respective reference velocity field components and pressure functions obtained using finite elements for \textbf{(a)}: $i = 1$ (25 fibers),  \textbf{(b)} $i = 2$ (36 fibers). The dashed squares indicate the porous zones $\mathcal{Z}_{P_{i}}$ in the mesoscale geometries.}
    \label{fig:2D geometries}
\end{figure}

In this section, we describe the benchmarks used for testing the proposed physics-informed methods. For $i=1,2$, we set $\mathcal{Z}_{i}=(0,1)^{2}$ with the ``textile rovings" $\mathcal{Z}_{P_{i}} = (0.28, 0.72)^{2}$ as the mesoscale geometries. We formally fix the number of ``fibers" for our microscale geometries $\mathcal{V}_{\mathcal{Z}_{1}}:=\mathcal{F}_{\downarrow}(\mathcal{Z}_{1})$ (25 fibers) and $\mathcal{V}_{\mathcal{Z}_{2}}:=\mathcal{F}_{\downarrow}(\mathcal{Z}_{2})$ (36 fibers) to resolve $\mathcal{Z}_{P_{1}}$ and $\mathcal{Z}_{P_{2}}$. These fibers correspond to domain perforations with radius $r_{1}=2.75 \times 10^{-2} $ in $\mathcal{V}_{\mathcal{Z}_{1}}$ and $r_{2}=2.5 \times 10^{-2}$ in $\mathcal{V}_{\mathcal{Z}_{2}}$, with no-slip boundary conditions enforced. We note that these are the normalized fiber sizes in the normalized to 1 cells $\mathcal{V}_{\mathcal{Z}_{i}}$. For periodic fiber arrangements in $\mathcal{V}_{\mathcal{Z}_{i}}$, $\boldsymbol{K}^{p}[\mathcal{V}_{\mathcal{Z}_{i}}]$ satisfies $\boldsymbol{K}^{p}[\mathcal{V}_{\mathcal{Z}_{i}}] = \boldsymbol{K}_{11}^{p}[\mathcal{V}_{\mathcal{Z}_{i}}] \mathcal{I}$, where $\mathcal{I} \in \mathbb{R}^{2\times 2}$ is the identity matrix and $\boldsymbol{K}_{11}^{p}[\mathcal{V}_{\mathcal{Z}_{i}}]$ is given by:
\begin{align}\label{permeability periodic}
\boldsymbol{K}_{11}^{p}[\mathcal{V}_{\mathcal{Z}_{i}}] = \frac{1}{\mu}\boldsymbol{PD}^{-1} \boldsymbol{U},
\end{align}
where $\boldsymbol{U}: = \boldsymbol{U}_{11}$ and $\boldsymbol{PD}: =\boldsymbol{PD}_{11}$ and $\mu = 1$ according to \eqref{averaging process}. Therefore, one problem \eqref{Stokes equation} is solved for $\boldsymbol{u}:=\boldsymbol{u}^{(1)}$, $p:=p^{(1)}$ and $\boldsymbol{f} := \boldsymbol{f}^{(1)} = 10\boldsymbol{e}^1$  using Taylor-Hood finite elements \cite{boffi2013mixed}. For this, we use the following boundary conditions:
\begin{align}\label{boundary conditions}
\boldsymbol{u} \ \   \text{periodic over } \  \mathcal{V}_{\mathcal{Z}_{i}}, \quad  \restr{p}{\Gamma_{\text{up}}} = \restr{p}{\Gamma_{\text{down}}},   \quad    p  = 0  \ \  \text{on} \ \  \Gamma_{\text{in}} \cup \Gamma_{\text{out}}, 
\end{align}
where $\Gamma_{\text{in}}$ and $\Gamma_{\text{out}}$ are the inlet (left) and outlet (right) boundaries of $\mathcal{V}_{\mathcal{Z}_{i}}$, and $\Gamma_{\text{up}}$ and $\Gamma_{\text{down}}$ are the top and bottom walls of $\mathcal{V}_{\mathcal{Z}_{i}}$, respectively. We computed dimensionless reference permeabilities $\boldsymbol{K}^{p}_{11} = (2.37 \pm 0.068) \times 10^{-4}$ for $\mathcal{V}_{1}$ and $\boldsymbol{K}_{11}^{p}  = (9.08 \pm 0.11) \times 10^{-5}$ for $\mathcal{V}_{2}$ using \eqref{permeability periodic}. To obtain the permeability measured in $\mathrm{m}^{2}$, the dimensionless permeability has to be rescaled by the square of a suitable characteristic length.  We perform averaging over $[0.3125 + l_{p}, 0.6875 - l_{p}]^2$ in \eqref{averaging process} with $l_{p} =0$, instead of over $\mathcal{V}_{\mathcal{Z}_{i}}^{F}$, to prevent overestimation of $\boldsymbol{K}^{p}_{11}$ caused by the flow outside the tows $\mathcal{F}_{\downarrow}(\mathcal{Z}_{P_{i}})$. The computation error is specified for variations of $l_{p} \in [0,0.045]$. Within this errors, the computed permeability values vary insignificantly with mesh refinements. For discretization of the Stokes-Brinkman equation with $\tilde{\mu} = \mu = 1$, we also applied Taylor-Hood finite elements. Our 2D microscale and mesoscale geometries, along with their respective finite element solutions, are shown in Fig.~\ref{fig:2D geometries}.

\subsection{Physics-informed neural networks as microscale surrogates}
\label{Section: Physics-informed neural networks}

Following \cite{raissi2019physics}, we aim to find a neural network $\widehat{NN}_{\theta, \psi}$  that learns the map from spatial coordinates to an approximation of the solution to \eqref{Stokes equation}: 
\begin{align} \label{PINN ansatz}
\boldsymbol{x} \mapsto \widehat{NN}_{\theta, \psi}(\boldsymbol{x}) := [\widehat{\boldsymbol{u}}_{\theta}(\boldsymbol{x}),  \ \widehat{p}_{\psi}(\boldsymbol{x})]. 
\end{align}
The parameters (weights and biases) $\theta \in \mathbb{R}^{n_{u}}$ and $\psi\in \mathbb{R}^{n_{p}}$ are obtained by minimizing the PINN objective:
\begin{equation}
\begin{aligned}\label{PINN loss}
\underset{\theta, \psi}{\min} \ \mathcal{J}(\theta, \psi): = \lambda^{r} \mathcal{J}^{r}(\theta, \psi)  +  \lambda^{b} \mathcal{J}^{b}(\theta)  + \lambda^{\text{div}} \mathcal{J}^{\text{div}}(\theta),  
\end{aligned}
\end{equation}
where positive constants $\lambda^{r}$, $\lambda^{\text{div}}$, $\lambda^{b}$, which weight the residual, divergence-free, and no-slip boundary penalties, respectively:
\begin{equation}\label{residual losses}
\begin{aligned}
\mathcal{J}^{r}(\theta, \psi) & = \frac{1}{N_{r}} \sum_{j=1}^{N_{r}} \big\lVert \frac{\partial \widehat{p}_{\psi}}{\partial x_{i}}(\boldsymbol{x}_{j}^{r}) - \mu  \Delta \widehat{\boldsymbol{u}}_{\theta} (\boldsymbol{x}_{j}^{r})  - \boldsymbol{f}\big \rVert_{2}^{2}, \\ 
\mathcal{J}^{\text{div}}(\theta)  & = \frac{1}{N_{r}} \sum_{j=1}^{N_{r}} \big| \nabla \cdot  \widehat{\boldsymbol{u}}_{\theta}(\boldsymbol{x}_{j}^{r})\big|^{2},  \quad  \mathcal{J}^{b}(\theta)  = \frac{1}{N_{b}} \sum_{j=1}^{N_{b}} \big \lVert \widehat{\boldsymbol{u}}_{\theta} (\boldsymbol{x}_{j}^{b}) \big \rVert_{2}^{2}, 
\end{aligned}
\end{equation}
where $\lVert  \cdot  \rVert_{2}$ is the Euclidean norm, the \textbf{collocation points} $\{\boldsymbol{x}_{j}^{r}\}_{j=1}^{N_{r}} \subset \mathcal{V}_{\mathcal{Z}}^{F}$, $\{\boldsymbol{x}_{j}^{b}\}_{j=1}^{N_{b}} \subset \partial \mathcal{V}_{\mathcal{Z}}^{S}$ in \eqref{residual losses} are sampled uniformly at random from $\mathcal{V}_{\mathcal{Z}}$ and its solid fiber surface $\partial \mathcal{V}_{\mathcal{Z}}^{S}$, and the spatial derivatives are computed using automatic differentiation, as it is usual for PINNs.

The adaptive Fourier feature network is employed for neural network-based approximation of $\widehat{\boldsymbol{u}}_{\theta}$; cf. \cite{wang2023expert}. The network follows the ansatz:
\begin{align}\label{nn ansatz}
f_{NN}^{u}(\boldsymbol{x}; \theta) = \boldsymbol{W}^{L_{f}}(\phi_{L_{f}-1} \circ \phi_{L_{f}-2}...\circ \phi_{L_{1}} \circ \phi_{E} )(\boldsymbol{x}) + \boldsymbol{b}^{L_{f}}. 
\end{align}
For $1 \leq l \leq L_{f}-1$ in \eqref{nn ansatz}, the $l$-th hidden layer is defined as: 
\begin{align*}
\phi_{L_{l}}(\boldsymbol{h}^{l-1}) & =\boldsymbol{h}^{l}: = \sigma (\boldsymbol{W}^{l}\boldsymbol{h}^{l-1} + \boldsymbol{b}^{l})
\end{align*}
where  $\boldsymbol{W}^{l} \in \mathbb{R}^{n_{l} \times n_{l-1}}$ and $\boldsymbol{b}^{l} \in \mathbb{R}^{n_{l}}$ are trainable weights and biases, $\sigma: \mathbb{R}^{n_{l}} \rightarrow \mathbb{R}^{n_{l}}$ are activation functions (we use $\tanh(\boldsymbol{x})$ by default), and the hidden states are $\boldsymbol{h}^{l} \in \mathbb{R}^{n_{l}}$, with $\boldsymbol{h}^{0}:=\phi_{E}(\boldsymbol{x})=[ \sin(2 \pi \boldsymbol{E}\boldsymbol{x}), \  \cos(2 \pi \boldsymbol{E}\boldsymbol{x})] \in \mathbb{R}^{d_{E}\times d}$. Choosing large $d_{E}$ in the embeddings $\phi_{E}$ helps mitigate the spectral bias \cite{rahaman2019spectral} toward learning low-frequencies by better capturing high-frequencies of the solution \cite{tancik2020fourier, wang2021eigenvector}. These high-frequency components are important, as they contribute significantly to the flow within the tows at the microscale and, as a result, affect the permeability computation. Trainable $\boldsymbol{E}\in \mathbb{R}^{\frac{d_{E}}{2} \times d}$ is beneficial as the spectral properties of \eqref{Stokes equation} are not known a priori. The network $f_{NN}^{p}(\boldsymbol{x}; \psi)$ for $\widehat{p}_{\psi}(\boldsymbol{x})$, with the same architecture as for $\widehat{\boldsymbol{u}}_{\theta}$, is also constructed. 

Periodic boundary conditions in \eqref{boundary conditions} for $\widehat{\boldsymbol{u}}_{\theta}$ can be enforced into the architecture using periodic embeddings as the first layer of the neural network:
\begin{align}
\mathrm{v}_{u}(\boldsymbol{x}) &= [\cos (2 \pi x_{1}), \  \sin (2 \pi x_{1}),  \ \cos (2 \pi x_{2}), \  \sin (2 \pi x_{2})], 
\end{align}
Similarly, for $\widehat{p}_{\psi}$, one uses $\mathrm{v}_{p}(\boldsymbol{x}) = [\cos (2 \pi x_{2}), \  \sin (2 \pi x_{2})]$. Periodicity can also be enforced via a soft penalization approach, but this introduces multiple penalty terms into the objective, significantly increasing optimization difficulty. The Dirichlet condition $p  = 0$ on $\Gamma_{\text{in}}$ and  $\Gamma_{\text{out}}$ is imposed penalty-free by modifying \eqref{nn ansatz} as follows: $[\widehat{NN}_{\theta}(\boldsymbol{x})]_{i} = g_{i}(\boldsymbol{x}) + s_{i}(\boldsymbol{x})[f_{NN}(\boldsymbol{x}; \theta)]_{i}$, where $g_{1}(\boldsymbol{x}) = 0$ and $s_{1}(\boldsymbol{x})=1$ for $\widehat{\boldsymbol{u}}_{\theta}$, and $g_{2}=0$, $s_{2}=4x_{1}x_{2}(1-x_{2})$ for $\widehat{p}_{\psi}$. As will be shown in our numerical experiments, applying PINNs with periodic boundary conditions yields unreliable results, likely because such conditions provide very limited information, leaving the network to rely primarily on residual penalty losses during training. In contrast, our dual-scale hybrid solver further informs the PINN loss with mesoscale information, enabling robust enforcement of periodic boundary conditions as well.

\subsection{Hybrid physics-informed neural network based dual-scale solver}
\label{Section: Hybrid solver approach}
\begin{figure}[H]
    \centering
    \includegraphics[width=0.8\textwidth]{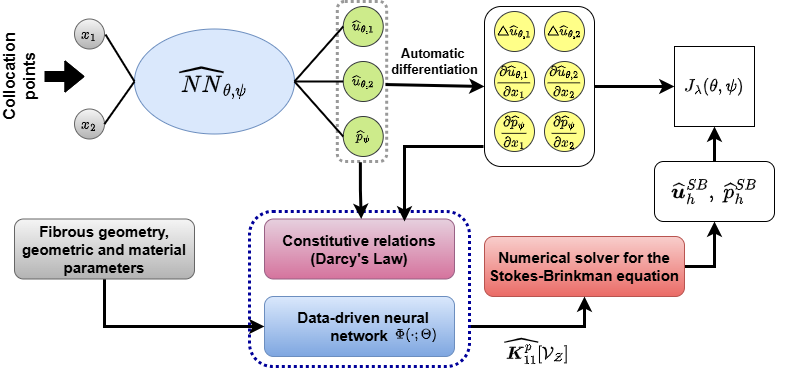}
    \caption{The schematic workflow for the hybrid neural network based dual scale solver.}
    \label{fig:hybrid solver}
\end{figure}

In our hybrid dual-scale approach, we exploit the dual-scale structure of textile stacks by using PINNs to simulate fluid flow at the microscale within fiber tows, while coupling the neural network solver with a mesoscale numerical solver that accurately approximates the flow outside the fiber tows. To enable the dual-scale coupling, we introduce a special term (coarse-scale regularizer), dependent on the solution to the Stokes-Brinkman equation, into the PINN objective \eqref{PINN loss}, thereby bridging the two scales:
\begin{align}\label{coarse-scale prior}
\mathcal{R}(\theta, \psi)  =  \frac{\lambda^{u}}{N_{u}} \sum_{j=1}^{N_{u}}  \big \lVert  \widehat{\boldsymbol{u}}_{\theta}(\boldsymbol{x}^{u}_{j}) - \widehat{\boldsymbol{u}}^{SB}(\boldsymbol{x}^{u}_{j}) \big \lVert_{2}^{2}   + \frac{\lambda^{p}}{N_{p}} \sum_{j=1}^{N_{p}}  | \widehat{p}_{\psi}(\boldsymbol{x}^{p}_{j}) - \widehat{p}^{SB}(\boldsymbol{x}^{p}_{j})|^{2} ,
\end{align}
where $\{\boldsymbol{x}_{j}^{u}\}_{j=1}^{N_{u}}, \{\boldsymbol{x}_{j}^{p}\}_{j=1}^{N_{p}} \subset \mathcal{Z}_{F}$, and $\lambda_{i}^{SB}>0$ are the weights. To evaluate \eqref{coarse-scale prior}, we need the PINN ansatz \eqref{PINN ansatz}, as well as the following learning-informed Stokes-Brinkman equation must be solved for $\widehat{\boldsymbol{u}}^{SB} =[\boldsymbol{u}^{SB}_{1},  \ \boldsymbol{u}^{SB}_{2}]$ and $\widehat{p}^{SB}$:
\begin{equation}\label{Stokes-Brinkman equation}
\begin{aligned}
- \tilde{\mu} \Delta \widehat{\boldsymbol{u}}^{SB} + \mu \big(\widehat{\boldsymbol{K}}_{\text{Me}}[\mathcal{V}_{\mathcal{Z}}] \big)^{-1} \widehat{\boldsymbol{u}}^{SB}  + \nabla \widehat{p}^{SB} &=  \boldsymbol{f}  \quad &\text{in} \ \mathcal{Z}, \\ 
\nabla \cdot \widehat{\boldsymbol{u}}^{SB} & =  0   \quad &\text{in} \ \mathcal{Z},
\end{aligned}
\end{equation}
where $\widehat{\boldsymbol{K}}_{\text{Me}}[\mathcal{V}_{\mathcal{Z}}] = \chi_{\mathcal{Z}^{F}} \cdot \infty + \chi_{\mathcal{Z}^{P}} \cdot \widehat{\boldsymbol{K}_{11}^{p}}[\mathcal{V}_{\mathcal{Z}}]$ and $\widehat{\boldsymbol{K}_{11}^{p}}[\mathcal{V}_{\mathcal{Z}}]$ is computed via \eqref{permeability periodic} using the outputs of the PINN ansatz \eqref{PINN ansatz}. Combining the PINN loss \eqref{PINN loss} with the coupling term \eqref{coarse-scale prior}, our hybrid solver problem reads:
\begin{align} \label{hybrid problem}
 \underset{\theta, \ \psi}{\text{min}}  \  J_{\boldsymbol{\lambda}}(\theta, \psi):= \mathcal{J}(\theta, \psi)  +  \mathcal{R}(\theta, \psi) \quad \text{subject to the constraints} \ \eqref{Stokes-Brinkman equation}.
\end{align}
Evaluating $J_{\boldsymbol{\lambda}}$ requires numerically solving \eqref{Stokes-Brinkman equation}. Since coarse-scale geometries are easier to mesh in general, we apply a standard numerical solver at the mesoscale. Enabling \eqref{Stokes-Brinkman equation} with the boundary conditions \eqref{boundary conditions} and using Taylor-Hood finite elements, we  approximate $\widehat{\boldsymbol{u}}^{SB}$ and  $\widehat{p}^{SB}$ with $\widehat{\boldsymbol{u}}^{SB}_{h}$ and $\widehat{p}^{SB}_{h}$, where $h>0$ denotes mesh resolution. To update \eqref{coarse-scale prior}, we numerically solve \eqref{Stokes-Brinkman equation} every $T$ iterations. The dual-scale solver structure is shown in Fig.~\ref{fig:hybrid solver} and is realized through optimization.  We note that a good initial value \( \widetilde{\boldsymbol{K}_{11}^{p}}[\mathcal{V}_{\mathcal{Z}}] \) for \( \widehat{\boldsymbol{K}_{11}^{p}}[\mathcal{V}_{\mathcal{Z}}] \) helps accelerate convergence of the mesoscale component in the hybrid solver. This guess can be obtained from a data-driven surrogate model $\Phi(\cdot, \Theta)$ with parameters $\Theta$. Such a surrogate can be constructed based on the principles described in Section \ref{Section: Methods for dual scale permeability prediction}. The hybrid approach then refines $\widetilde{\boldsymbol{K}_{11}^{p}}[\mathcal{V}_{\mathcal{Z}}]$ through an iterative process guided by the PINN loss.

The hybrid objective \eqref{hybrid problem} is optimized using a variant of the gradient descent algorithm with the generic form
\begin{align}\label{discrete gradient flow}
\begin{bmatrix}
\theta_{k+1} \\
\psi_{k+1} 
\end{bmatrix} = \begin{bmatrix}
\theta_{k} \\
\psi_{k} 
\end{bmatrix} - l_{k} \begin{bmatrix}
\nabla_{\theta} \mathcal{J}(\theta_{k}, \psi_{k} ) + \nabla_{\theta}\mathcal{R}(\theta_{k}, \psi_{k})\\ 
\nabla_{\psi} \mathcal{J}(\theta_{k}, \psi_{k} ) + \nabla_{\psi}\mathcal{R}(\theta_{k}, \psi_{k})
\end{bmatrix}, 
\end{align}
where $l_{k}>0$ is the learning rate. In the discrete gradient flow \eqref{discrete gradient flow}, variations in the magnitudes of the involved gradients are observed. Indeed, comparing the histograms of the back-propagated gradients reveals that the entries of $\nabla_{\theta} \mathcal{J}^{b}(\theta)$ are concentrated near zero, while the residual gradients $\nabla_{\theta} \mathcal{J}^{r}(\theta, \psi)$ span much wider ranges. This is a well-known phenomenon of unbalanced gradients, which causes the gradient flow \eqref{discrete gradient flow} to be stiff, leading to convergence issues \cite{wang2021understanding}. In our case, stiffness increases with the number of fibers. To partially mitigate this, we apply a weight scaling technique to $\boldsymbol{\lambda} := (\lambda^{r}, \lambda^{\text{div}}, \lambda^{b}, \lambda^{u}, \lambda^{p})$, based on the gradient magnitudes of the respective loss terms, as suggested in \cite{wang2023expert}. After iteration $k_{c}$, we apply exponential annealing to the weights in \eqref{coarse-scale prior} to reduce the impact of microscale–mesoscale flow discrepancies on the microscale approximation. The hybrid optimization, outlined in Algorithm \ref{alg:hybrid solver}, requires initializing $\theta_{0}$, typically using Glorot initialization \cite{glorot2010understanding} is applied. Algorithm~\ref{alg:hybrid solver} also requires computing the projection:
\begin{align}\label{permeability projection}
\mathcal{P}_{[\boldsymbol{K}_{\text{LB}}^{p}, \boldsymbol{K}_{\text{UB}}^{p}]}\big(\widehat{\boldsymbol{K}_{11}^{p}}[\mathcal{V}_{\mathcal{Z}}]\big) = \max \{\boldsymbol{K}_{\text{LB}}^{p}[\mathcal{V}_{\mathcal{Z}}], \min \{ \widehat{\boldsymbol{K}_{11}^{p}}[\mathcal{V}_{\mathcal{Z}}], \  \boldsymbol{K}_{\text{UB}}^{p}[\mathcal{V}_{\mathcal{Z}}]\} \},
\end{align}
where $\boldsymbol{K}_{\text{LB}}^{p}[\mathcal{V}_{\mathcal{Z}}]$ and $\boldsymbol{K}_{\text{UB}}^{p}[\mathcal{V}_{\mathcal{Z}}]$ denote the lower and upper bounds for the predicted permeability $\widehat{\boldsymbol{K}_{11}^{p}}[\mathcal{V}_{\mathcal{Z}}]$, respectively. \eqref{permeability projection} ensures that $\widehat{\boldsymbol{K}_{11}^{p}}[\mathcal{V}_{\mathcal{Z}}]$ remains in between $\boldsymbol{K}_{\text{LB}}^{p}[\mathcal{V}_{\mathcal{Z}}]$ and $\boldsymbol{K}_{\text{UB}}^{p}[\mathcal{V}_{\mathcal{Z}}]$, thereby preventing poor PINN outputs by regularizing our learning process with a well-informed coarse-scale solution.

\begin{algorithm}[ht!]
\caption{Hybrid physics-informed dual-scale solver}
\label{alg:hybrid solver}
  \textbf{Input:} $\widetilde{\boldsymbol{K}_{11}^{p}}[\mathcal{V}_{\mathcal{Z}}], \ \boldsymbol{K}_{\text{LB}}^{p}[\mathcal{V}_{\mathcal{Z}}], \  \boldsymbol{K}_{\text{UB}}^{p}[\mathcal{V}_{\mathcal{Z}}], \  \theta_{0}$, maximal iteration number $k_{\text{max}}$, $k_{c} \in \mathbb{N}$, $T \in \mathbb{N}$, $\gamma_{u},\gamma_{p} \in \mathbb{R}_{+}$ .  \\
  \textbf{Output:}   $(\widehat{\boldsymbol{u}}_{\theta},\widehat{p}_{\psi})$, $\widehat{\boldsymbol{K}_{11}^{p}}[\mathcal{V}_{\mathcal{Z}}]$, $(\widehat{\boldsymbol{u}}^{SB}_{h}, \widehat{p}^{SB}_{h})$ 
  \begin{algorithmic}[1]
  \WHILE {$0 \leq k \leq k_{\text{max}}-1$}
  \IF{$k\bmod{T} = 0$}
  \IF{$k \leq k_{c}$}
  \STATE {Update: $\boldsymbol{\lambda}_{k} \leftarrow \text{Gradient-based weight scaling}\big(\nabla_{\theta,  \psi} J_{\boldsymbol{\lambda}_{k}}(\theta_{k}, \psi_{k}),  \  \boldsymbol{\lambda}_{k} \big)$.}
  \ELSE 
  \STATE {Update: $\boldsymbol{\lambda}_{k}^{u}, \  \boldsymbol{\lambda}_{k}^{p}  \leftarrow  \exp{(\gamma_{u} (k_{c}-k))}\cdot \boldsymbol{\lambda}_{k_{c}}^{u},  \ \exp{(\gamma_{p} (k_{c}-k))}\cdot \boldsymbol{\lambda}_{k_{c}}^{p}$.}
  \ENDIF
  \STATE {Update: $\widehat{\boldsymbol{K}_{11}^{p}}[\mathcal{V}_{\mathcal{Z}}](\theta_{k}, \psi_{k})  \leftarrow  \text{Darcy's Law}\big(\widehat{NN}_{\theta_{k}, \psi_{k}}\big)$ according to \eqref{permeability periodic}}
  \STATE {Project: $\widehat{\boldsymbol{K}_{11}^{p}}[\mathcal{V}_{\mathcal{Z}}](\theta_{k}, \psi_{k})  \leftarrow  \mathcal{P}_{[\boldsymbol{K}_{\text{LB}}^{p}, \boldsymbol{K}_{\text{UB}}^{p}]}\big(\widehat{\boldsymbol{K}_{11}^{p}}[\mathcal{V}_{\mathcal{Z}}](\theta_{k}, \psi_{k})\big)$ using \eqref{permeability projection}}
  \STATE {Compute FEM solution to \eqref{Stokes-Brinkman equation} and update coupling term \eqref{coarse-scale prior}.}
  \ENDIF
  \STATE{Compute $\nabla_{\theta, \psi} J_{\boldsymbol{\lambda}_{k}}(\theta_{k}, \psi_{k})$ using automatic differentiation}
  \STATE {$\theta_{k+1}, \psi_{k+1} \leftarrow \text{Optimizer}\big(\nabla_{\theta, \psi} J_{\boldsymbol{\lambda}_{k}}(\theta_{k}, \psi_{k}), \ \text{optimizer hyperparameters} \big)$} 
  \STATE {$k \leftarrow k+1$}
  \ENDWHILE
  \end{algorithmic}
\end{algorithm}

\section{Comparison of dual-scale methods for permeability prediction} 
\label{Section: comparison of scale-bridging approaches}

As the hybrid dual-scale solver does not yet support 3D models with large amount of fibers, the results are split into two sections.

\subsection{Comparison of NUM, SBM and FRM for permeability prediction}

\begin{figure}[H]
    \centering
    \includegraphics[width=0.9\linewidth]{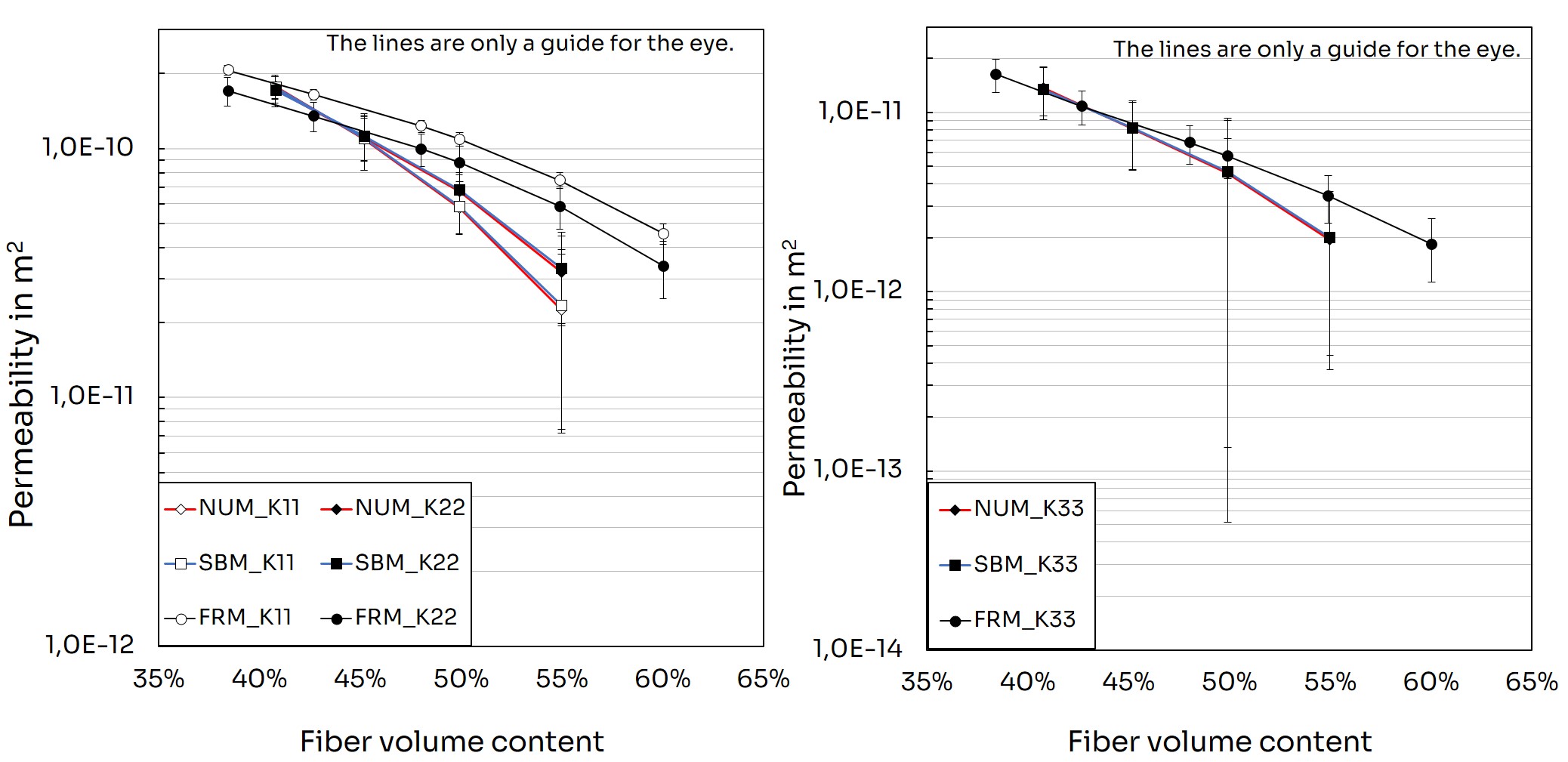}
    \caption{Diagrams of the in-plane (left) and out-of-plane (right) permeability results plotted over FVC of NUM, SBM and FRM}
    \label{fig:5}
\end{figure}

This section compares the permeability predictions obtained from the numerical upscaling method (NUM), the scale-bridging method using data-driven machine learning (SBM), and the fully resolved model (FRM). The diagrams and data in Table \ref{tab:result_overview} and Fig. \ref{fig:5} illustrate the permeability values \(K_{11}\), \(K_{22}\), and \(K_{33}\) as functions of the fiber volume content (FVC).

A clear trend of decreasing permeability with increasing FVC is evident across all directions, which is attributed to reduced pore space and increased fiber packing density. For the in-plane permeabilities \(K_{11}\) and \(K_{22}\), NUM and SBM show very close agreement across the entire investigated FVC range and exhibit good consistency with the FRM predictions, especially at lower FVC levels. The standard deviations (SD) and coefficients of variation (CV) remain moderate in this range, with CV values typically between 10\% and 25\%.

In the out-of-plane direction \(K_{33}\), both NUM and SBM tend to underestimate \(K_{33}\) compared to FRM, particularly at higher FVC levels. For example, at 50\% FVC, CV values for \(K_{33}\) reach up to 97.1\% (NUM20) and 98.9\% (SBM20), indicating greater prediction uncertainty. This behavior reflects the increased sensitivity of through-thickness flow to microscale structural variations and the limitations of simplified permeability assumptions in this direction. These differences can be attributed to the randomized model generation and the resulting structural variations, which lead to significant differences in predicted permeability values.

In terms of computational effort, FRM requires substantially higher runtimes, ranging from approximately 88 to 120 hours depending on the FVC, due to its detailed resolution and large model sizes. By contrast, NUM and SBM complete simulations in under 11 hours (e.g., NUM15: 10.5 h, SBM15: 9.0 h), making them significantly more efficient. It should be noted that FRM simulations were executed on a high-performance computing cluster, whereas NUM and SBM were run on a workstation (Intel Core i9-14900K, 128 GB RAM). Therefore, the runtimes are not directly comparable.

Despite minor differences in permeability values, particularly for \(K_{33}\), both NUM and SBM reproduce similar overall trends and maintain consistent in-plane predictions compared to FRM. The segment-wise assignment of microscale permeability in SBM does not lead to notable accuracy improvements over NUM but slightly increases runtime in some cases. It is assumed that the high-fidelity FRM approach provides the most accurate results since it involves minimal simplifications and no loss of information due to scale separation. The SBM method accounts for structural variability on the microscale and the resulting local permeability variations more precisely than NUM, thus approximating the FRM results while requiring significantly less computing time. However, for this particular study, the role of FRM results as reference must be critically examined, due to the reductions in model size and resolution that were necessary to allow computing.

All these methods have specific advantages and limitations, each aiming to efficiently and accurately capture the physical phenomena in dual-scale structures. The presented results highlight the trade-off between model fidelity and computational efficiency and provide a solid foundation for discussing the potential of hybrid approaches, including the implementation of a PINN-based hybrid solver, as shown in the following section.

\subsection{Comparison of physics-informed methods for permeability prediction}

For our experiments, we use the benchmarks from Section \ref{Section: Benchmark models for physics-informed approaches}. For the neural networks, we consider the following setup: the architecture from Section \ref{Section: Physics-informed neural networks}, which includes a Fourier feature layer with $d_{E} = 256$ and three hidden layers, each with 128 neurons, is used for both $\widehat{\boldsymbol{u}}_{\theta}$ and $\widehat{p}_{\psi}(\boldsymbol{x})$. We use the Adam optimizer \cite{kingma2014adam} with an exponential learning rate schedule: the initial learning rate is set to $l_{0} = 1 \times 10^{-3}$, with a decay rate of 0.9. From the finite element discretization of \eqref{Stokes-Brinkman equation}, mesh points $\{\boldsymbol{x}_{j}^{c}\}_{j=1}^{N_{SB}} \subset \mathcal{Z}_{F} \cap \mathcal{V}_{\mathcal{Z}}^c$ are used in \eqref{coarse-scale prior}, with $\mathcal{V}_{\mathcal{Z}_{i}} \setminus \mathcal{V}_{\mathcal{Z}}^c$, where $\mathcal{V}_{\mathcal{Z}}^c = [0.22, 0.78]^2$. However, mesh points within $0.015$ of the periodic boundary in \eqref{boundary conditions} are excluded from \eqref{coarse-scale prior} to avoid the influence of the coarse-scale solution on learning the periodic boundary conditions.  From $\mathcal{V}_{\mathcal{Z}_{1}}$,  $N_r = 60800$ collocation points are sampled: $15000$ from $\mathcal{V}_{\mathcal{Z}}^c$ and $45000$ from $\mathcal{V}_{\mathcal{Z}_{1}} \setminus \mathcal{V}_{\mathcal{Z}_{1}}^c$, plus $200$ points on each of $\Gamma_{\text{in}}$, $\Gamma_{\text{out}}$, $\Gamma_{\text{up}}$ and $\Gamma_{\text{down}}$ to support periodic embeddings $\mathrm{v}_{u}$ and $\mathrm{v}_{p}$. Additionally, $200 \times 25$ collocation points per fiber are used, totaling $N_b = 5000$ in \eqref{residual losses}. From $\mathcal{V}_{\mathcal{Z}_{2}}$, $N_r = 60800$ collocation points are sampled: $25000$ from $\mathcal{V}_{\mathcal{Z}}^c$ and $70000$ from $\mathcal{V}_{\mathcal{Z}_{2}} \setminus \mathcal{V}_{\mathcal{Z}}^c$, along with periodic boundary points as in $\mathcal{V}_{\mathcal{Z}_{1}}$, and $200 \times 36$ collocation points per fiber. We train using a full set of collocation points, without exploiting the distributed optimization frameworks based on batch training with stochastic optimization algorithms; cf. \cite{stiller2020large}. For Algorithm \ref{alg:hybrid solver}, we used $k_{\text{max}}=25000$, $k_{c}=5000$, $T=250$, $\gamma_{u},\gamma_{p} =2.5 \times 10^{-4}$, $\widetilde{\boldsymbol{K}_{11}^{p}}[\mathcal{V}_{\mathcal{Z}}] = 4.5 \times 10^{-4}, \ \boldsymbol{K}_{\text{LB}}^{p}[\mathcal{V}_{\mathcal{Z}}] = 5 \times 10^{-5}, \  \boldsymbol{K}_{\text{UB}}^{p}[\mathcal{V}_{\mathcal{Z}}] = 5 \times 10^{-4}$.

\begin{figure}[H]
    \centering
    \includegraphics[width=0.9\linewidth]{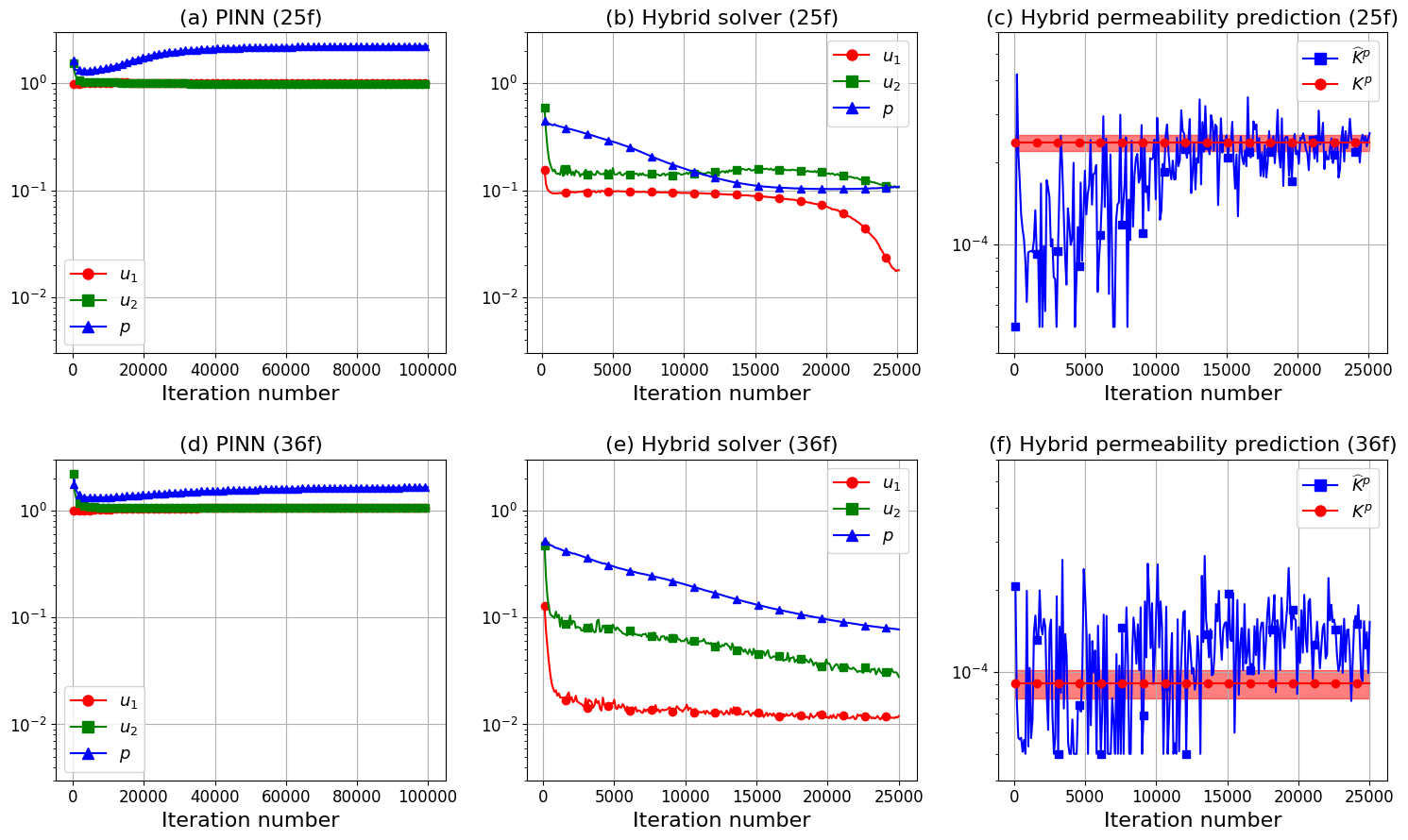}
    \caption{Relative $l_{2}$ errors for $u_{1}, u_{2}$ and $p$ vs iterations of PINNs (a,d) and hybrid solver (b,e). Abbreviations `25f" and `36f" stand for $\mathcal{V}_{\mathcal{Z}_{1}}$ and $\mathcal{V}_{\mathcal{Z}_{2}}$, respectively. Permeability prediction $\widehat{\boldsymbol{K}}^{p}$ (blue line) vs iterations of the hybrid solver in $\mathcal{V}_{\mathcal{Z}_{1}}$(c) and in $\mathcal{V}_{\mathcal{Z}_{2}}$(f). The red band in (c) and (f) represents the error in the computed reference permeability $\boldsymbol{K}^{p}$, indicated by the constant red line.}
    \label{fig:error curves}
\end{figure}

We first note that with the above setup, we were unable to reach any reasonable accuracy with PINNs. For both geometries $\mathcal{V}_{\mathcal{Z}_{1}}$ and $\mathcal{V}_{\mathcal{Z}_{2}}$, the errors for both velocity field and the pressure stagnate at the very beginning and do not show any further progress after 100k iterations, see Fig. \ref{fig:error curves}(a) and Fig. \ref{fig:error curves}(d), respectively. We attribute these difficulties to the neural network’s inability to learn periodic boundary conditions in an early training phase: without correctly enforcing periodicity, identifying the PDE solution becomes impossible, and the resulting initial instabilities are further amplified by the stiff gradient flow induced by the penalty term enforcing the no-slip boundary condition, making the learning process highly challenging without appropriate regularization, such as the coarse-scale regularization in the hybrid approach. 

\begin{figure}[H]
    \centering
    \includegraphics[width=0.85\linewidth]{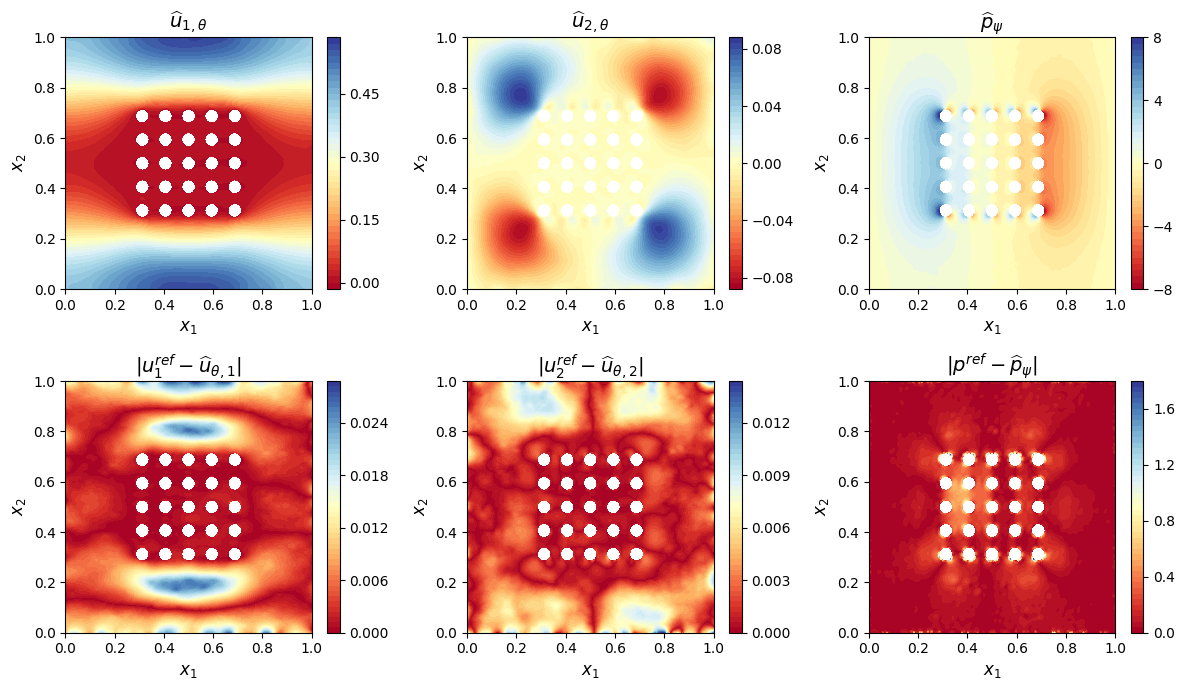}
    \caption{Hybrid solver results for the microscale flow and pointwise errors (relative to the Stokes reference solution computed with finite elements) for  $\mathcal{V}_{\mathcal{Z}_{1}}$ .}
    \label{fig:pointwise errors 25}
\end{figure}

\begin{figure}[H]
    \centering
    \includegraphics[width=0.85\linewidth]{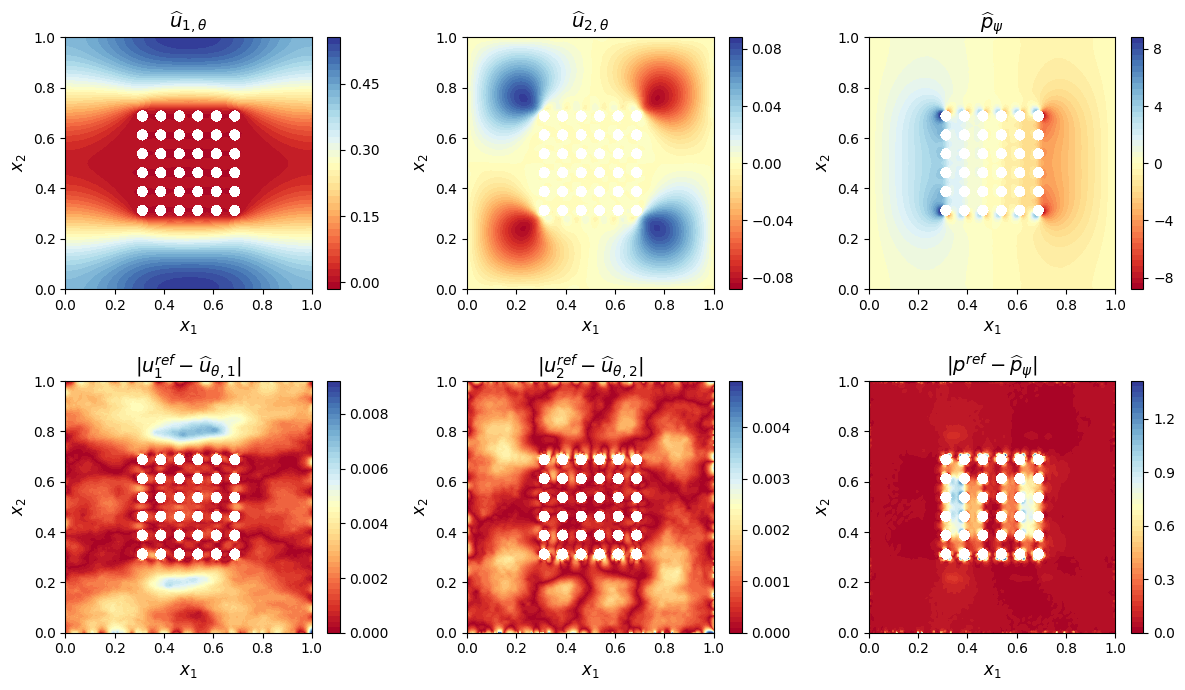}
    \caption{Hybrid solver results for the microscale flow and pointwise errors (relative to the Stokes reference solution computed with finite elements) for $\mathcal{V}_{\mathcal{Z}_{2}}$.}
    \label{fig:pointwise errors 36}
\end{figure}

\begin{figure}[H]
    \centering
    \includegraphics[width=0.8\linewidth]{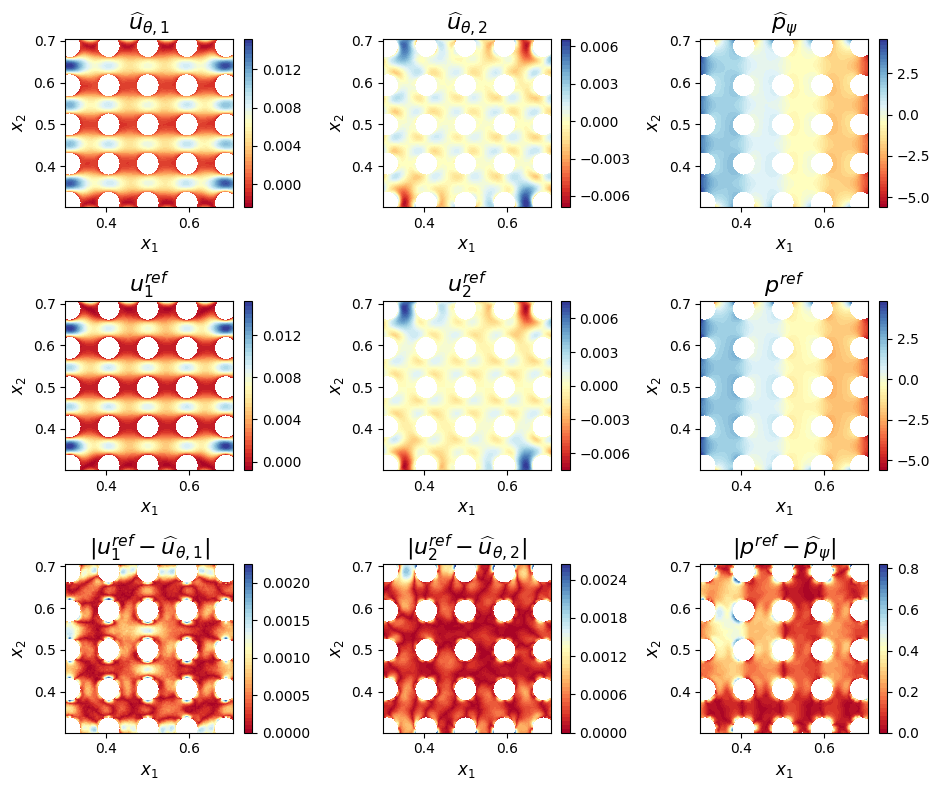}
    \caption{Hybrid solver results for $\mathcal{V}_{\mathcal{Z}_{1}}$: flow components and pressure in the fiber tow.}
    \label{fig:microstructure 25}
\end{figure}

\begin{figure}[H]
    \centering
    \includegraphics[width=0.8\linewidth]{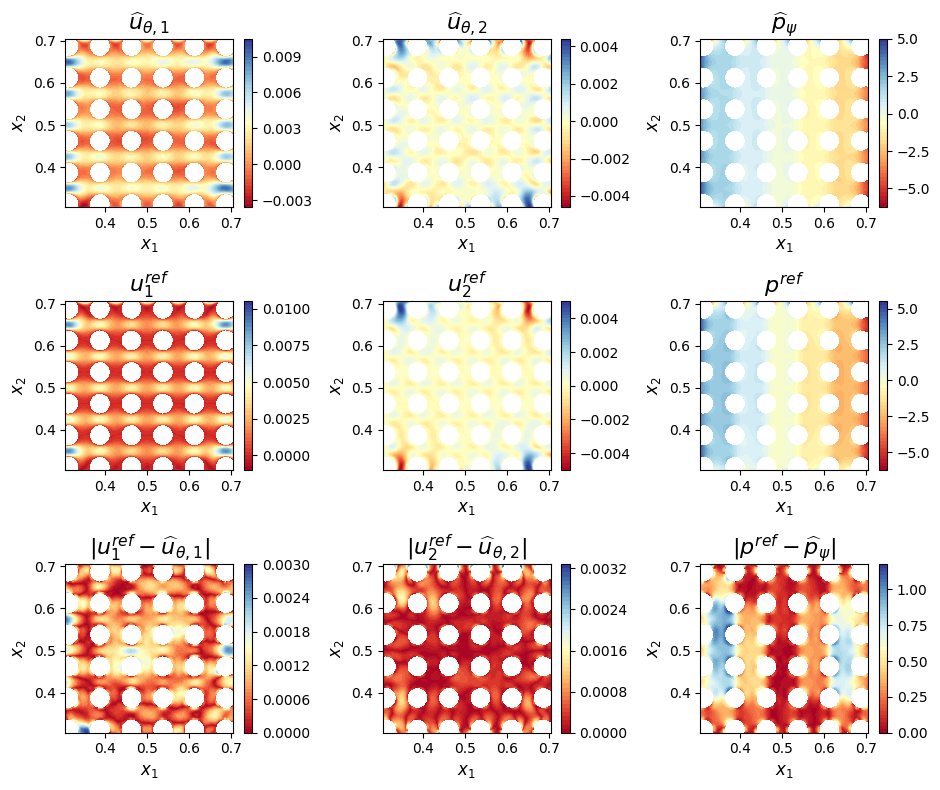}
    \caption{Hybrid solver results for $\mathcal{V}_{\mathcal{Z}_{2}}$: flow components and pressure in the fiber tow.}
    \label{fig:microstructure 36}
\end{figure}

Indeed, compared to the standard PINNs, the hybrid approach yields reasonable accuracy  within 25k iterations with respect to the finite element reference solutions, see Fig. \ref{fig:error curves}(b) and Fig. \ref{fig:error curves}(e). For $\mathcal{V}_{\mathcal{Z}_1}$, we get the $l^{2}$ errors of 1.84\text{E}{-02} and 1.07\text{E}{-01} for $\widehat{u}_{1, \theta}$, and $\widehat{u}_{2, \theta}$, respectively, and
$1.08\text{E}{-01}$ for $\widehat{p}_{\psi}$. For $\mathcal{V}_{\mathcal{Z}_2}$, we get the $l^{2}$ errors of 1.23\text{E}{-02} and 2.82\text{E}{-02} for $\widehat{u}_{1, \theta}$ and $\widehat{u}_{2, \theta}$ , respectively, and
$8.12\text{E}{-02}$ for $\widehat{p}_{\psi}$. We can see that Fig. \ref{fig:pointwise errors 25} and Fig. \ref{fig:pointwise errors 36} show good agreement with the reference solutions, and the resulting approximations are even better for the second geometry. This can be attributed to a better match between the mesoscale and microscale models, thereby yielding more stable training. However, this improvement in convergence is primarily limited to the low-frequency components that capture the flow outside the tows, which are well informed by the mesoscale model. The hybrid solver is also capable of capturing complex fluid flow with significant high-frequency content within the tows for both geometries, as shown in Fig. \ref{fig:microstructure 25} and Fig. \ref{fig:microstructure 36}, which is essential for computing permeability. Still, accurately approximating the refined details of the microscale flow within the tow of $\mathcal{V}_{\mathcal{Z}_{2}}$ is more challenging, due to increased gradient flow stiffness in the optimization due to the larger number of fibers. This is also evidenced by the overestimation of predicted permeability and its instability in Fig. \ref{fig:error curves}(f). In contrast, despite some minor oscillations, there is good agreement between the predicted and reference permeability for $\mathcal{V}_{\mathcal{Z}_{1}}$, as shown in Fig. \ref{fig:error curves}(c).

The current PINN-based approach is significantly limited by its training speed — a well-known issue in the scientific community \cite{mcgreivy2024weak}. The hybrid approach takes 20 to 25 minutes to train (on a single A100 GPU with 40 GB RAM) for $\mathcal{V}_{\mathcal{Z}_{1}}$ and $\mathcal{V}_{\mathcal{Z}_{2}}$, respectively. However, there is evidence \cite{stiller2020large} of an almost linear speedup in PINN training time with the number of GPUs, provided the algorithm supports highly distributed implementation. Substantial mathematical and algorithmic development is required, and a more powerful GPU cluster must be employed to scale the approach to complex applications such as 3D fibrous geometries.

\section{Conclusion}

This work presents several approaches for predicting the permeability of mesoscale textile models, based on various permeability determination methods: NUM, SBM, FRM and PINNs-based approach. The NUM approach represents the current state of the art. The SBM improves upon NUM by accounting for the influence of microscale permeability on the mesoscale behavior. Despite its increased complexity, SBM maintains computational efficiency through the use of data-driven surrogates. Although permeability values vary across methods, as noted in \cite{syerko2023benchmark, syerko2024mesoscale}, SBM shows potential to enhance both efficiency and accuracy. In contrast, FRM is theoretically the most accurate method due to its fully resolved modeling approach but faces prohibitive computational costs, emphasizing the need for efficient scale-bridging methods or the integration of hybrid deep learning frameworks to mitigate these limitations.
The SBM approach does not provide substantial accuracy improvements over NUM but accounts for microscale structural variability more explicitly. Nevertheless, both methods maintain consistent in-plane predictions and offer a practical balance for large-scale simulations where the high computational costs of FRM are prohibitive. While NUM and SBM significantly reduce runtime requirements (below 11 hours), FRM demands runtimes up to 120 hours due to its detailed resolution and large model sizes.

In our physics-informed contribution, we compared the capabilities of standard PINNs with our novel two-scale hybrid PINN approach for approximating microscale flow within fibers using 2D geometries. We applied periodic boundary conditions by embedding periodicity into the neural network architecture. Despite incorporating Fourier features and gradient-based loss balancing, as recommended in best practices, standard PINNs failed to accurately approximate the flow. We attribute this to the limited guidance provided by periodic boundary conditions during training, highlighting the need for additional sources of information. The hybrid dual-scale solver with the same neural network architecture successfully enforces periodic boundary conditions by incorporating additional information from a mesoscale numerical solver, while also successfully capturing high-frequency flow details in the tows. Both of these advances are important for accurate permeability estimation using physics-informed approaches and, consequently, enable effective scale bridging with PINNs. However, reducing the computational cost and runtime of our physics-informed approach remains key to practical application. Scaling to such scenarios will require many collocation points and large networks, posing  further computational challenges that must be addressed through (hybrid) domain decomposition techniques combined with efficient (hybrid) optimization algorithms.

Overall, we believe that our hybrid dual-scale framework is a first step toward integrating diverse methods for scale bridging, while capturing the reciprocal dependencies across scales. This approach may be particularly useful for complex composite material simulations, such as in unsaturated flow scenarios. Addressing all these challenges remains a central focus of our future work.


\section{Acknowledgments}
All authors acknowledge the support of the Leibniz Collaborative Excellence Cluster under project  ML4Sim (funding reference: K377/2021).






\bibliographystyle{elsarticle-num} 
\bibliography{bib.bib}






\newpage

\begin{sidewaystable}[htbp]
\centering
\footnotesize
\renewcommand{\arraystretch}{1.2}

\begin{tabular}{|l|cc|ccc|ccc|ccc|c|}
\hline
\textbf{Model} & \multicolumn{2}{c|}{\textbf{FVC}} & \multicolumn{3}{c|}{\textbf{K11 in m\textsuperscript{2}}} & \multicolumn{3}{c|}{\textbf{K22 in m\textsuperscript{2}}} & \multicolumn{3}{c|}{\textbf{K33 in m\textsuperscript{2}}} & \textbf{RT in h} \\
 & MV & SD & MV & SD & CV & MV & SD & CV & MV & SD & CV & MV \\
\hline
NUM30 & 40,8\% & 1,96\% & 1,78E-10 & 1,93E-11 & 10,8\% & 1,73E-10 & 2,12E-11 & 12,2\% & 1,37E-11 & 4,10E-12 & 30,0\% & 5,9 \\
SBM30 & 40,8\% & 1,96\% & 1,76E-10 & 1,87E-11 & 10,6\% & 1,71E-10 & 2,38E-11 & 13,9\% & 1,35E-11 & 4,41E-12 & 32,7\% & 4,9 \\
NUM25 & 45,2\% & 2,17\% & 1,09E-10 & 2,69E-11 & 24,7\% & 1,11E-10 & 2,14E-11 & 19,4\% & 8,11E-12 & 3,33E-12 & 41,1\% & 5,9 \\
SBM25 & 45,2\% & 2,17\% & 1,10E-10 & 2,79E-11 & 25,5\% & 1,12E-10 & 2,21E-11 & 19,8\% & 8,19E-12 & 3,45E-12 & 42,1\% & 6,3 \\
NUM20 & 49,9\% & 2,40\% & 5,76E-11 & 1,24E-11 & 21,5\% & 6,73E-11 & 1,13E-11 & 16,8\% & 4,59E-12 & 4,46E-12 & 97,1\% & 8,4 \\
SBM20 & 49,9\% & 2,40\% & 5,86E-11 & 1,30E-11 & 22,1\% & 6,84E-11 & 1,18E-11 & 17,3\% & 4,68E-12 & 4,63E-12 & 98,9\% & 7,9 \\
NUM15 & 55,0\% & 2,63\% & 2,25E-11 & 1,53E-11 & 68,0\% & 3,20E-11 & 1,26E-11 & 39,3\% & 1,95E-12 & 1,51E-12 & 77,4\% & 10,5 \\
SBM15 & 55,0\% & 2,63\% & 2,34E-11 & 1,60E-11 & 68,3\% & 3,29E-11 & 1,31E-11 & 39,7\% & 2,00E-12 & 1,63E-12 & 81,6\% & 9,0 \\
FRM10 & 38,4\% & 0,09\% & 2,06E-10 & 8,80E-12 & 4,3\% & 1,70E-10 & 2,20E-11 & 12,9\% & 1,64E-11 & 3,50E-12 & 21,4\% & 119,5 \\
FRM20 & 42,7\% & 0,10\% & 1,65E-10 & 7,99E-12 & 4,9\% & 1,35E-10 & 1,86E-11 & 13,7\% & 1,08E-11 & 2,35E-12 & 21,6\% & 98,5 \\
FRM30 & 48,0\% & 0,11\% & 1,23E-10 & 6,74E-12 & 5,5\% & 9,97E-11 & 1,51E-11 & 15,1\% & 6,78E-12 & 1,63E-12 & 24,0\% & 96,2 \\
FRM33 & 49,9\% & 0,12\% & 1,09E-10 & 6,55E-12 & 6,0\% & 8,80E-11 & 1,41E-11 & 16,0\% & 5,70E-12 & 1,42E-12 & 25,0\% & 88,4 \\
FRM40 & 54,9\% & 0,13\% & 7,46E-11 & 5,33E-12 & 7,1\% & 5,87E-11 & 1,13E-11 & 19,2\% & 3,43E-12 & 1,02E-12 & 29,9\% & 97,6 \\
FRM46 & 60,0\% & 0,14\% & 4,55E-11 & 4,26E-12 & 9,4\% & 3,37E-11 & 8,80E-12 & 26,1\% & 1,84E-12 & 7,08E-13 & 38,5\% & 107,5 \\
\hline
\end{tabular}
\vspace{2mm}
\footnotesize

MV = mean value, SD = standard deviation, CV = coefficient of variation, RT = runtime of the flow simulation.

\caption{Results of NUM, SBM and FRM at different fiber volume contents}
\label{tab:result_overview}
\end{sidewaystable}

\end{document}